\documentclass{article}
\usepackage{iclr2015,times}
\usepackage{hyperref}
\usepackage{url}
\usepackage{amsmath}
\usepackage{amssymb}
\usepackage{graphicx}
\usepackage{amsmath}
\usepackage{amsfonts}
\newcommand*{\LargerCdot}{\raisebox{-0.25ex}{\scalebox{1.2}{$\cdot$}}}
\iclrfinalcopy
\title{Explorations on high dimensional landscapes}

\author{
Levent Sagun\textsuperscript{1}, V. U\u{g}ur G\"uney\textsuperscript{2}, G\'erard Ben Arous\textsuperscript{1}, \& Yann LeCun\textsuperscript{1,3} \\ 
\textsuperscript{1}Courant Institute of Mathematical Sciences, New York, NY 10012 \\
\textsuperscript{2}Department of Physics, The City University of New York, New York, NY 10016\\
\textsuperscript{3}Facebook AI Research, 770 Broadway, New York, NY 10003\\
\texttt{sagun@cims.nyu.edu}, \texttt{uguney@gc.cuny.edu} \\ 
\texttt{benarous@cims.nyu.edu}, \texttt{yann@cs.nyu.edu}\\
}

\begin{document}

\maketitle

\begin{abstract}

Finding minima of a real valued non-convex function over a high dimensional space is a major challenge in science. We provide evidence that some such functions that are defined on high dimensional domains have a narrow band of values whose pre-image contains the bulk of its critical points. This is in contrast with the low dimensional picture in which this band is wide. Our simulations agree with the previous theoretical work on spin glasses that proves the existence of such a band when the dimension of the domain tends to infinity. Furthermore our experiments on teacher-student networks with the MNIST dataset establish a similar phenomenon in deep networks. We finally observe that both the gradient descent and the stochastic gradient descent methods can reach this level within the same number of steps.

\end{abstract}

\section{Introduction}

Many problems of practical interest can be rephrased as forms of an optimization problem in which one needs to locate a point in a given domain that has some optimal energy or cost value. Given a dynamics on such a surface the question of where it will stop on the landscape on which it is moving is a challenging one, especially if the terrain is rugged. Complex\footnote{Even though we use `complex' to refer to systems that has exponentially many critical points, we would like to point out that to the best of our knowledge there is no agreed upon definition of what complex a system is across disciplines.} systems can be described by functions that are highly non-convex, that lives on high dimensional spaces and they have exponentially many critical points. 

One such problem arises in the context of learning: Given a set of input-output pairs, one seeks to find a function whose values roughly match the ones in the set and which also performs well on new inputs. That is, it learns the task described by the set it began with. To this end, a loss function is formed which is optimized over the parameters of the desired function. Performance of this loss minimization is tested against another set of input-output pairs.

The second problem we consider comes from statistical physics: In magnetic materials, the spins are aligned according to interactions with their neighbors. Over time equilibrium is reached when particles settle down to a configuration that has a reasonably low energy, if not the lowest possible. Stability of this state is tested against small fluctuations.

Some machine learning problems, such as deep networks, hint at similarity with spin systems of statistical mechanics. Finding connections between the two became an attractive research area. For a nice survey of theoretical results between Hopfield networks and Boltzmann machines, see \citet{Agliari2014}, and  \citet{1742-5468-2012-07-P07009}. In a slightly different approach \citet{NIPS2014_5486} and \citet{Choromanska} exhibit that stochastic gradient descent might end up in critical points other than local minima.

The energy landscapes of spin glasses are of interest in themselves. One of the main questions concerns the number of critical points of a given index at a given level. For a detailed analysis of this in spherical spin glasses, see \citet{Auffinger2010}, \citet{Auffinger2013} and \citet{Auffinger2014}. The first of this sequence of papers will be used in this present paper in the corresponding experimental section. It establishes the existence of an energy level, which we named \textit{floor}\footnote{Throughout this paper the energy levels that an algorithm can not pass in a reasonable amount of time will be referred to as \textit{floor}. For example in the spin glass context of the following section it corresponds to the smallest energy of the highest value of the function $\Theta(u)$ in equation \ref{Th}, which is $-E_{\infty}$.}, in which bulk of the low index critical points lie in the absence of an external field. It is proved that this level contains exponentially many local minima. For an algorithm to pass this level the search time should be exponentially long. Yet the floor lies just above the ground state, therefore from an optimization point of view it does not matter whether the algorithm reaches to the global minimum or local minimum at the floor. For a similar analysis in a slightly more general context using random matrices, see \citet{Fyodorov2013}, and \citet{Fyodorov2013a}. For a review of random matrices and extreme value statistics, see \citet{Dean2008}. The added external field as a tunable parameter allows changing the topology of the landscape. Inspired by this work, \citet{Mehta} gives a detailed experimental analysis of the case in which there are polynomially many critical points, and \citet{Mehta2} counts critical points for third order interactions in which there are exponentially many critical points. 

In this work, we focus on the case that has exponentially many critical points. In such high dimensional non-convex surfaces, a moving object is likely to get stuck in a local minimum point or a plateau of flat areas that is close to degenerate. A reasonable algorithm should have enough noise to make the particle jump out of those critical points that have high energy. However, if the landscape has exponentially many critical points that lie around the same energy level, then jumping out of this well will lead to another point of similar energy, clearly not resulting in any improvement. The existence of such a floor will increase the search time for points that have lower energy values regardless of the method of choice. Alternatively, one can attempt to change the system so that floor is at a different, more favorable value, namely closer to the global minimum. One will of course be curious about the performance of such a point in the learning context, or its stability in the spin glass context. 

So we ask two questions:

1) Where does the particle end up and how good is that point for the task at hand?

2) What do various descent algorithms do differently?

We perform experiments in spin glasses and in deep networks to address those questions above. The differences in the graph connectivities of spins and weights indicate that the two systems are not mathematically analogous. Rather we propose that they are special cases of a more general phenomenon which is not yet discovered. The authors hope to extract further knowledge from such complex systems that will shed light to a thorough theoretical understanding in the future.

\section{Setting for experiments and previous theoretical work}

\subsection{Mean field spherical spin glass model}

The simplest model of a magnetic material is the one in which atoms have spins with two states: +1 for spin up, or -1 for spin down. Considering pairwise interactions (in other words 2-body interactions) between neighbors gives the total energy of the system: $-\sum_{i j} w_i w_j $ where $w$ represents spin sign of the particle located at position $i$. Mean field assumption ignores the lattice geometry of particles and assumes all interactions have equal strength. Introducing a weak external field leads all particles to favor alignment, which incidentally corresponds to the minimum energy configuration which will be achieved at the point where all of the particles have spin up (or down for that matter). This is a rather simple landscape whose global minimum is easy to achieve. This assumes no frustration, that is, all particles favor alignment. This model is known as the Currie-Weiss model. If we have an alloy in which some particle pairs favor alignment and some favor dis-alignment, the picture changes drastically. This introduces a whole new set of constraints which may not be simultaneously attained, leading to glassy states. One way to model this is to introduce random interaction coefficients between particles: $\sum_{i j} x_{i j}w_i w_j $ this model, which has a rather complex energy landscape, is known as the Sherrington-Kirkpatrick model. See \citet{Agliari2014} and \citet{Sherrington2014} for a survey on spin glasses and their connection to complex systems.

We can also disregard the discrete states and put the points on the sphere and consider $3$-body interactions of $N$ particles. Taking a point $w\in S^{N-1}(\sqrt{N}) \subset \mathbb{R}^N$ and $x_{(\LargerCdot) } \sim \text{ Gaussian}(0,1)$, the Hamiltonian with standard normal interactions is given by $\sum_{i j k} x_{i j k}w_i w_j w_k $. We have arrived at the model of interest for the purposes of this paper. While this model has been studied previously in great detail, our particular focus is on the location of critical points. The main results we will use in this section and further details on the model can be found in \citet{Auffinger2010} and references therein. 

\citet{Auffinger2010} considers a more general version that involves any combinations of interactions not only 2 or 3. The main reason we consider triplets is because it is the smallest system that has exponentially many critical points. Before moving any further we would like to remark on why there are only polynomially many critical points in the binary interaction case. For $\sum w^2_i=N$ and $x_{ij} \sim$ Gaussian(0,1) the Hamiltonian of the system is given by
\[
H(w)=\sum_{i \sim j}w_iw_jx_{ij}=(Mx,x)
\]
which is a quadratic form where $M_{ij}=\frac{x_{ij}+x_{ji}}{2}$ is a random $N\times N$ Gaussian Orthogonal Ensemble matrix. This system has $2N$ critical points at eigenvectors and their corresponding energy values are eigenvalues scaled by $N$. 
 
\subsubsection{Floor of the Hamiltonian}

For $w\in S^{N-1}(\sqrt{N}) \subset \mathbb{R}^N$ and $x_{(\LargerCdot) } \sim \text{ Gaussian}(0,1)$ consider the function
\begin{equation}
 H_{N}(w)=\frac{1}{N}\sum_{i, j, k}^Nx_{ijk}w_{i}w_{j}w_{k} \label{ham}
\end{equation}
Once the random coefficients are chosen, the landscape that will be explored is fixed. Note that as a continuous function on a compact set $H_{N}$ attains its minimum and maximum. Also, since the landscape is symmetric through origin, its local or global maximum points have similar energy values with the opposite sign. We only focus on its minimum.

Let $\mathcal{N}_{N,k}(u)$ denote the total number of index-$k$ critical points of $H_N$ that lies below the level $Nu$. In other words $\mathcal{N}_{N,k}(u)$ is the random variable that counts critical points in the set $\{w: H_N(w)\leq Nu\}$. \citet{Auffinger2010} finds asymptotic expected value of this quantity in logarithmic scale:
\begin{equation}
  \mathbb{E}[\mathcal{N}_{N,k}(u)] \asymp e^{N\Theta_k(u)} \label{power}
\end{equation}

\begin{figure}[h]
\centering
\includegraphics[scale=0.3]{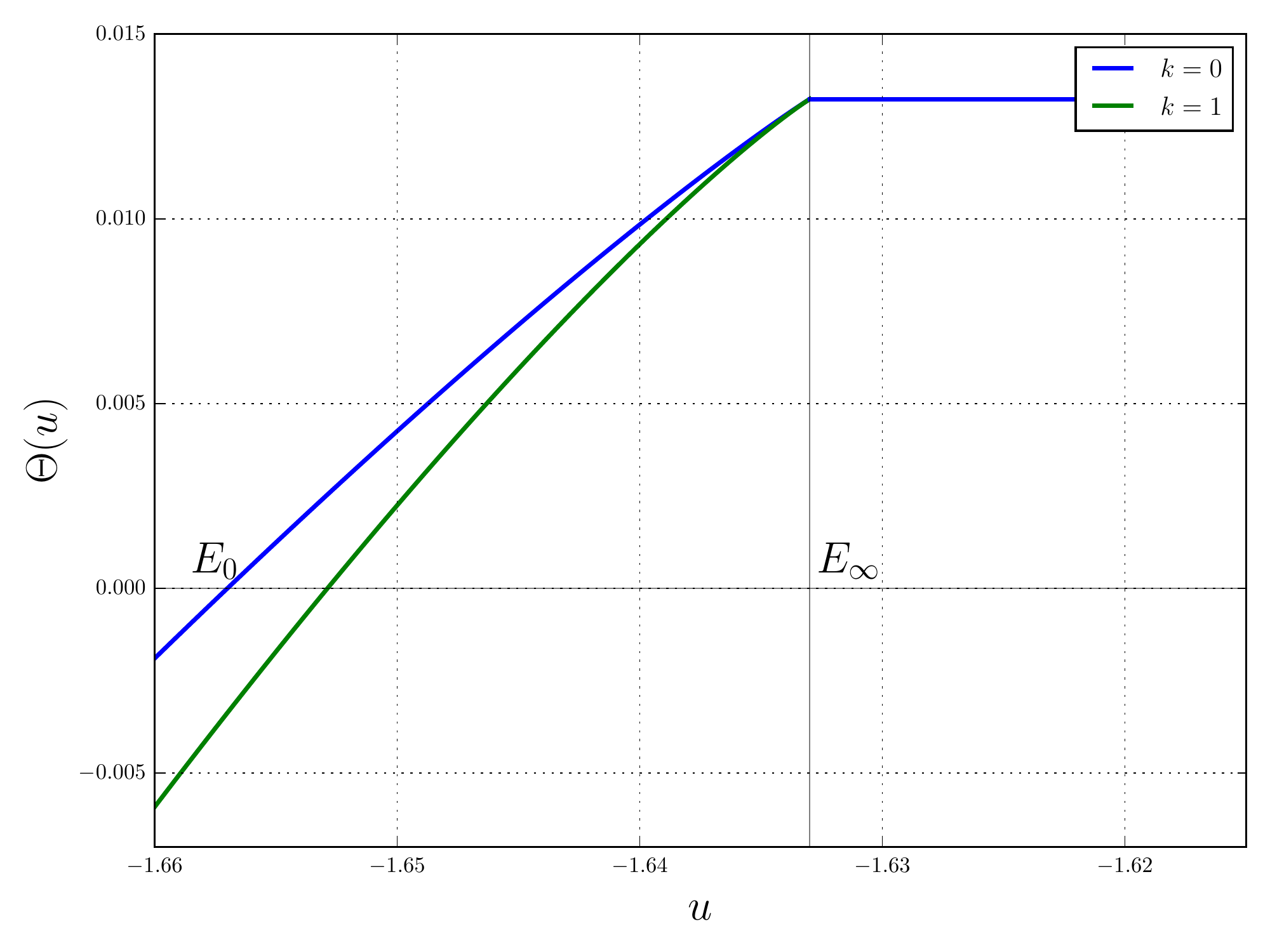}
\caption{Plot of $\Theta_0(u)$ (upper curve for local minima) and $\Theta_1(u)$ (lower curve for index 1). Their values agree at the point $-E_{\infty}$, which is the asymptotic value of the floor, and remain constant. Also note that the energy values on the horizontal axis are scaled by $N$.} \label{Th}
\end{figure}

In other words $\Theta(u)$ is a measure of the cumulative number of critical points from the ground state to the level $Nu$. An exact analytic description of the function $\Theta(u)$ can be found in \citet{Auffinger2010}. Figure \ref{Th} shows that analytic expression for $k=0$, local minima, and $k=1$, saddles of index 1. It shows, as expected, that $\Theta$ in equation \eqref{power} is non-decreasing as it counts the number of critical points cumulatively. $\Theta$ becomes positive\footnote{Note that in the binary interaction case described above, the exponent in \eqref{power} never crosses beyond zero otherwise it would imply exponentially many critical points. So that the binary interaction case, or the two-body case is an example of a simple system.}, and keeps increasing until it becomes constant at a value denoted as $-E_{\infty}$, which is the lowest $u$ for which $\Theta$ is at its maximum value. Hence the number of critical points of low index do not increase at high energy values. 

Clearly the function in \eqref{ham} is a Gaussian random variable as it is formed by a sum of independent Gaussians. The $1/N$ normalization factor is chosen for the model so that $\mathrm{Var}(H) = N$. This implies that we have a random field with mean zero and variance $N$ at each point. Another implication of the size of variance of the Hamiltonian is that its extensive quantities scale with $N$, for which \citet{Auffinger2010} gives, 

\begin{equation}
\liminf_{N\rightarrow \infty} \big(\min_w \frac{H_{N}(w)}{N}\big) \geq -E_0 \label{GS}
\end{equation}

where $-E_0$ is the zero of $\Theta$ function. In words, the global minimum of the Hamiltonian on the sphere is bounded from below by $-NE_0$.  This lower bound gives a measure on how far a given point is away from the ground state. For our experiments the energy at the ground state and floor is calculated to be $-N1.657$ and $-N1.633$, respectively. For practical purposes the floor level is close to the ground state. At the level of $-N1.633$ (the vertical line in Figure 1) the landscape is expected to exponentially many points that are local minimum or saddles of low index, and the exponent is at its maximum. Therefore probability of finding a local minimum that lies below the floor level is exponentially small. This leads us to conjecture that this floor is the first place to get stuck when using a descent method which is confirmed in our simulations. Diving deeper would require a long search time. Therefore trying to find points that has lower energy levels is not necessary and even realizable. 

\subsection{Setting for MNIST}

Consider a probability measure $\mu$ on $\mathbb{R}^n$, centralized, that represents data, and a function $\mathcal{G}$ from $\mathbb{R}^n$ to $\mathbb{R}$ that correctly labels data. True loss then measures how well a given function, say $G(w)$, approximates $\mathcal{G}$:
\[
\begin{aligned}
d(w)&=d(\mathcal{G},G(w)) \\
&=\int_{\mathbb{R}^n}d(\mathcal{G}(x),G(w)(x))d\mu(x) \\
\end{aligned}
\] 

For an integer $P$ consider i.i.d. training sample, $x^p$ for $p=1,...,P$, of the measure $\mu$ and define the empirical training loss:
\[
 \mathcal{L}_{\text{Train}}(w) = \frac{1}{P} \sum_{p=1}^P L(x^p,w) 
\] 
where $L$ is the loss per sample which can be the mean square loss, the hinge loss or cross-entropy. Also by the law of large numbers: 
\[
\mathcal{L}_{\text{Train}}(w)\rightarrow E_{\mu}[L(x,w)] = d(w)  \text{ $\mu$-a.s. as } P \rightarrow \infty \text{.} 
\] 
The limit holds pointwise so that the sampled loss approximates well the true loss as the number of samples increases. Moreover, by the central limit theorem the following holds pointwise:
\[
\begin{aligned}
\sqrt{P}(\mathcal{L}_{\text{Train}}(w)-d(w)) &\xrightarrow{d.} \mathcal{N}(0,\sigma^2(w)) \\
\text{where   } \sigma^2(w) &= E_{\mu}[L^2(x,w)] - d^2(w) \\
\end{aligned}
\] 

So that the fluctuations of this approximation is pointwise Gaussian. In real life problems true loss is not accessible, instead we have $\mathcal{L}_{\text{Train}}(w)$ which approximates $d(w)$. From the above convergence properties we expect both the test and training loss to converge with good accuracy to the true loss. Therefore we expect $\mathcal{L}_{\text{Train}}(w) \sim \mathcal{L}_{\text{Test}}(w)$ as well if there is enough data for both functions. A natural question is how close are they to each other? If we sample points from the floor of training surface, do they necessarily correspond to the floor of the test surface? How does the smaller scale fluctuations effect learning? We address these questions in the rest of the paper.

\section{Simulations and experiments}

\begin{figure}[h]
\centering
\includegraphics[scale=0.38]{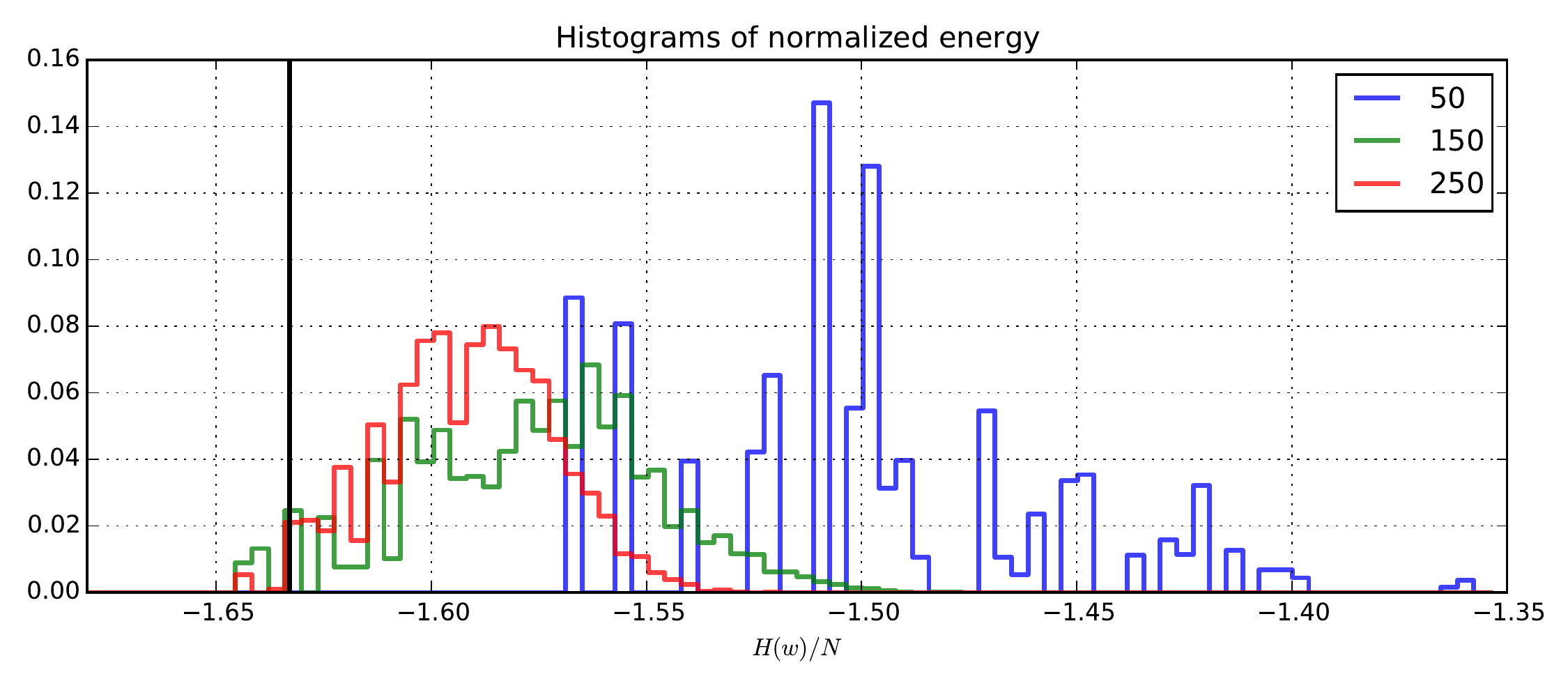}
\includegraphics[scale=0.38]{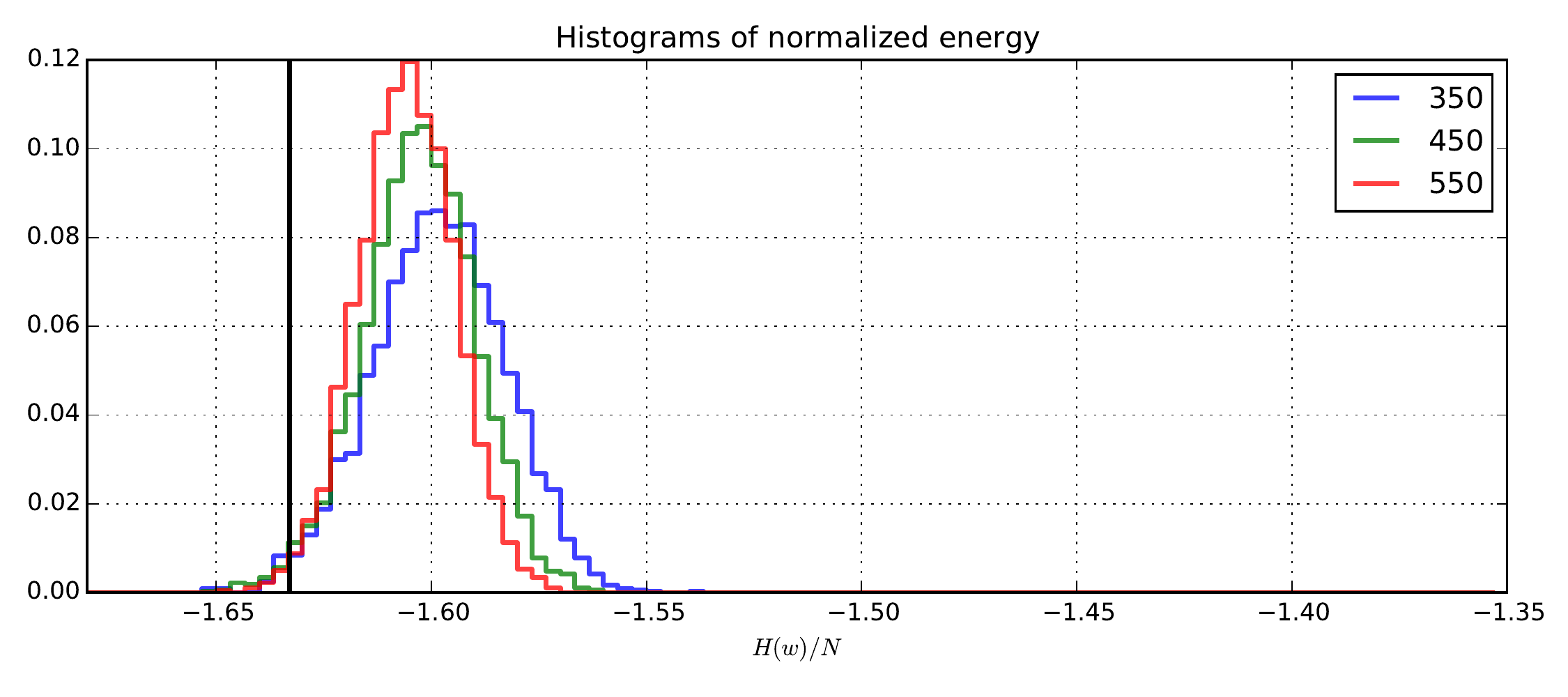}
\caption{Vertical black lines indicate the theoretical value of the floor in the large $N$ limit. When $N$ is small the system gets trapped at bad energy values (upper histogram). When $N$ is finite but large, floor is a narrow band (lower histogram).} \label{spinglass}
\end{figure}

Our simulations in the first part below show existence of a critical dimension before which the landscape is hectic and does not really show any sign of a well-defined value that algorithms converge. This implies that in low dimensions the surface is not trivial, and that there are traps at high energy levels and those traps are located at somewhat arbitrary values as seen in figure \ref{spinglass}. On the other hand high dimensional picture is drastically different. The landscape is trivial in the following sense: Starting from a random point, and following the gradient descent algorithm almost always leads to a very narrow band of values. Moreover in the second part we show that this descent is irrespective of which algorithm is being used. 

\subsection{Floor for high dimensional systems}

\subsubsection{Spin glasses}

Simulations of the spin glass model shows a clear qualitative difference between low dimensional and high dimensional surfaces. Namely, low dimensional surfaces do not exhibit a well defined floor however high dimensional surfaces does. Another important feature is that the floor level is very close to the global minimum, therefore floor level is enough for practical purposes of optimization, even when we set aside the problem of reaching global minimum.

The gradient descent algorithm is used on spin glass landscape. Procedure starts with fixing random couplings. Given the dimension $N$, sample $N^3$ many i.i.d. standard normal random variables: 
\begin{enumerate}
\item{Pick a random element $w$ of the sphere $S^{N-1}(\sqrt{N})$ as a starting point for each trial.}
\item{Iteratate\footnote {Note here the constraint that the gradient is tangential to the sphere.} $w^{t+1} = w^t - \gamma_t \nabla_wH(w^t)$, and normalize, $\sqrt{N}\frac{w^{t+1}}{||w^{t+1}||} \leftarrow w^{t+1} $.}
\item{Stop when the gradient size is below the order of $10^{-5}$.}
\end{enumerate} 

For each $N$ the experiment is repeated 5000 times. And the stopping criteria is the norm of the gradient vector. The theoretical results of the previous section holds true asymptotically. At this point, to the best of the knowledge of authors, there is no proof of how fast the energy of critical points concentrate around the limiting floor value. We hope this simulation will lead to a prediction in this direction. Nevertheless it is observed that the points found with this procedure lies within a narrow band of values that is just above the asymptotic floor level that is given in the previous section.

\subsubsection{Tri-partite spin glass: a toy model for multilayer Restricted Boltzmann Machines} 
\begin{figure}[h] 
\begin{center} 
\includegraphics[scale=0.2]{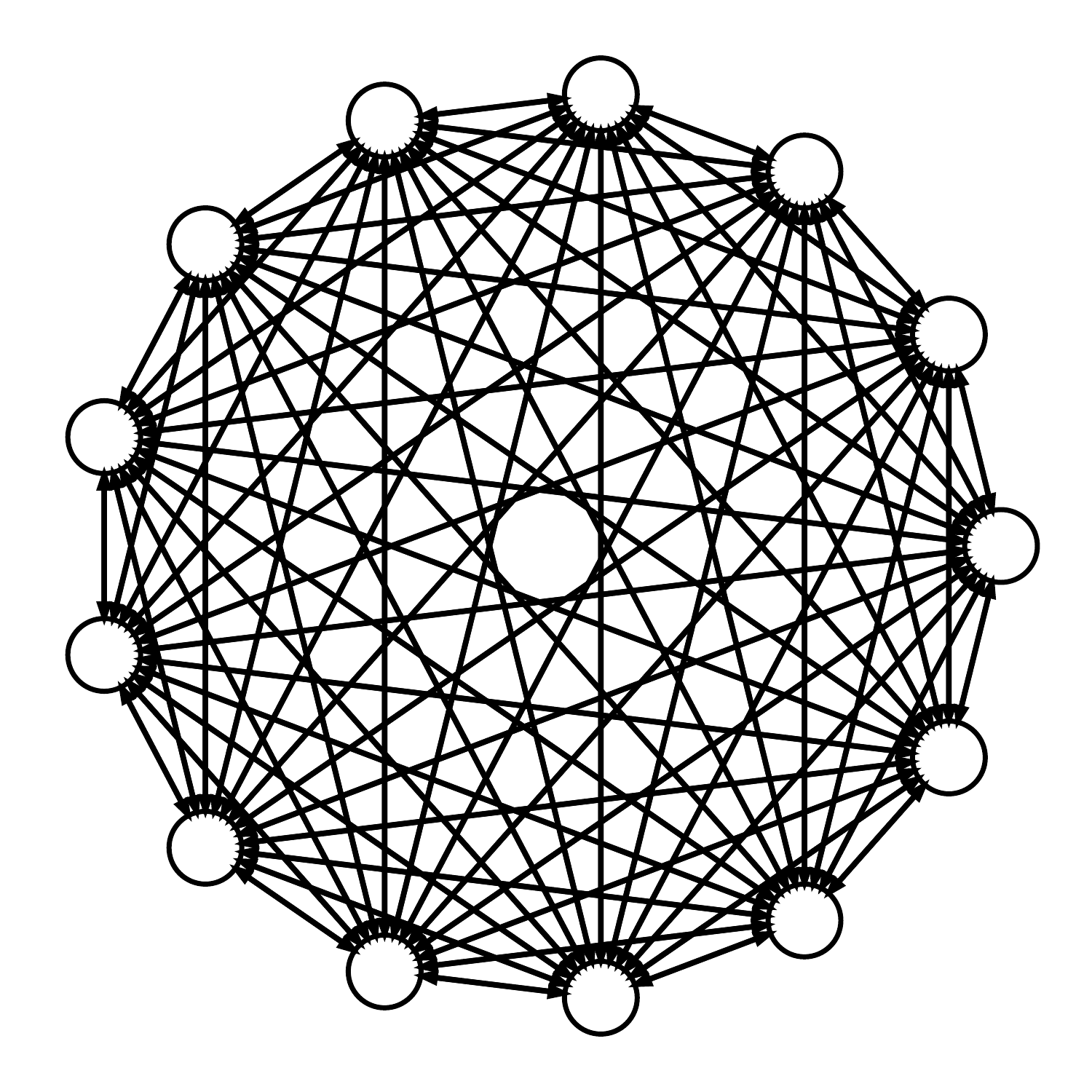} 
\hspace{1.5 cm}
\includegraphics[scale=0.2]{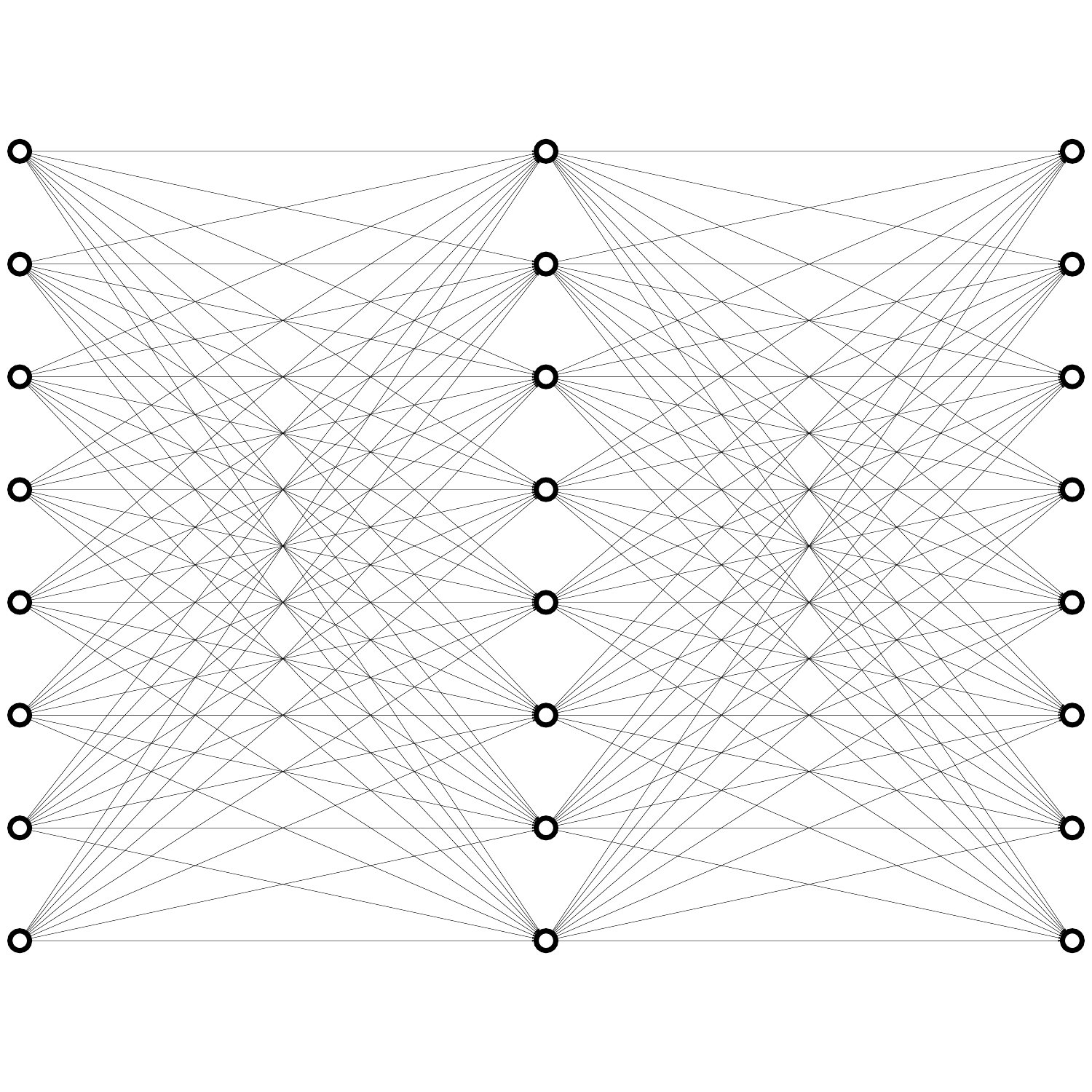} 
\end{center} 
\caption{Graph connectivity: p-spin spin glass vs tri-partite spin glass.} 
\end{figure} 
The graph structure of a fully connected spin glass above is highly coupled, in this section we modify the function so the terms of the polynomial have a layered structure. This is achieved by simulating the uncoupled version of the above spin glass model in a way that mimics the layered structure of a multilayer Restricted Boltzmann Machine. From an optimization point of view a similar consideration can be found in \citet{frieze_et_al:LIPIcs:2008:1752}. 
\begin{figure}[h]
\begin{center}
\includegraphics[scale=0.38]{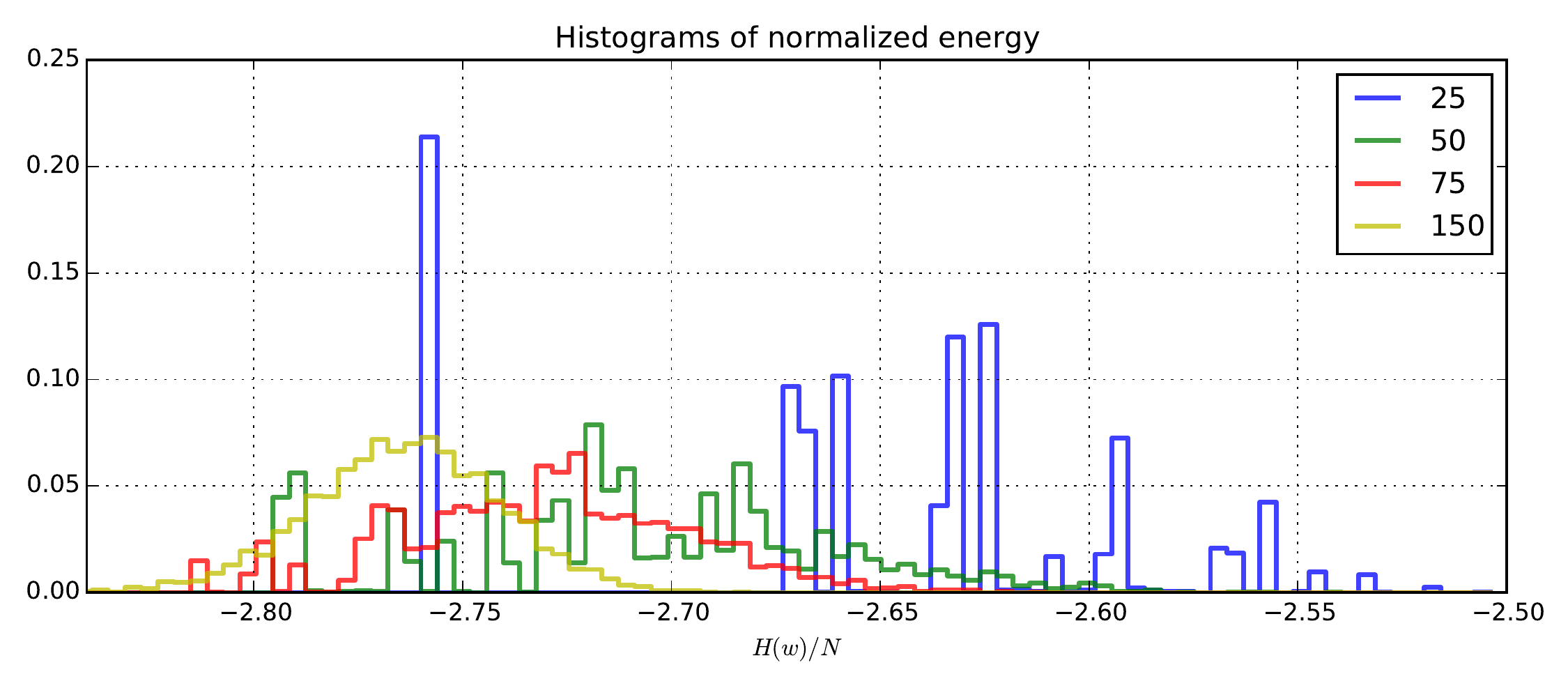} 
\includegraphics[scale=0.38]{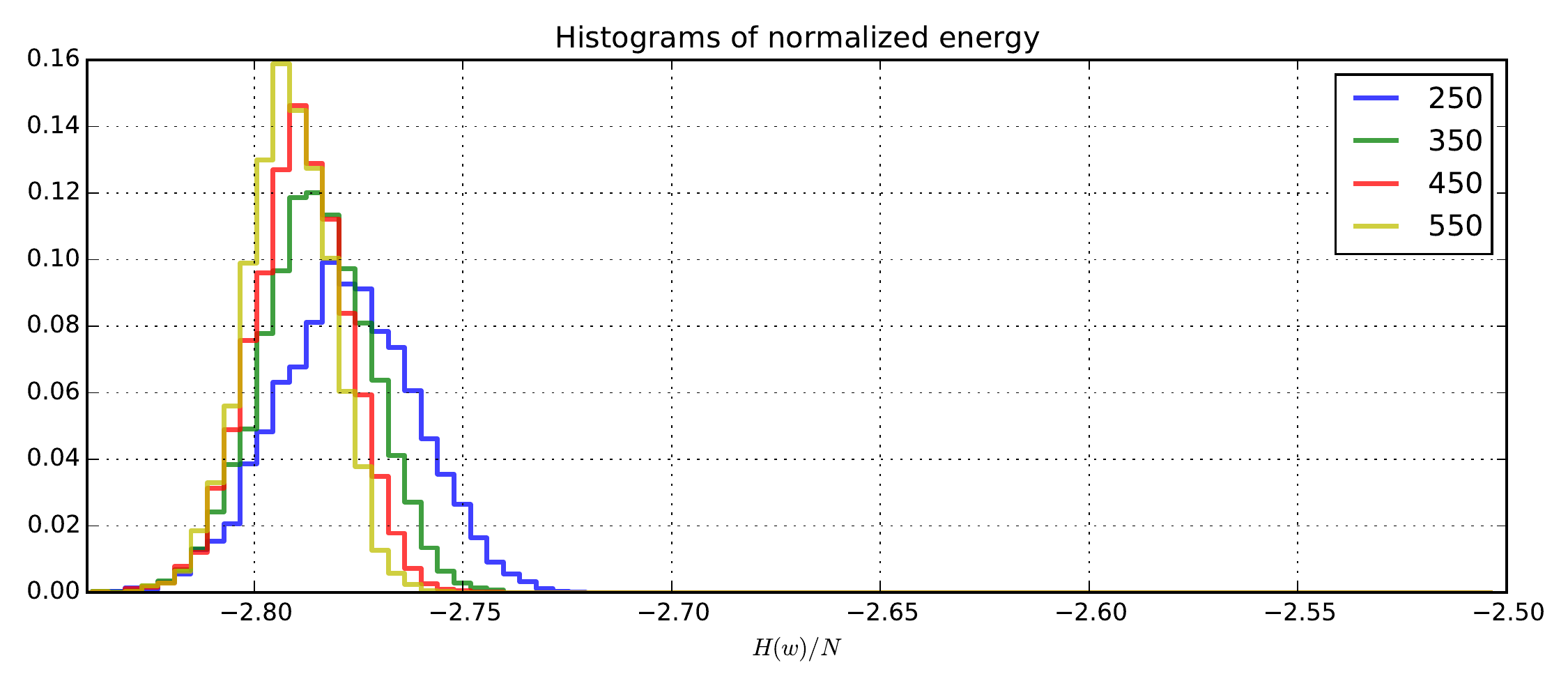}
\end{center}
\caption{Tri-partite spins low dimension vs high dimension comparison.} \label{uncoupled}
\end{figure}
In this new setting, take three different points on the sphere, or equivalently, take one point on the product of three spheres to form the tri-partite Hamiltonian. Instead of equation \ref{ham} the uncoupled function at hand becomes $\tilde{H}_{N}(w)=\frac{1}{N}\sum x_{ijk}w^1_{i}w^2_{j}w^3_{k}$. Then perform gradient descent over all parameters and normalize the resulting $3N$ dimensional vector on the product of spheres. Since there are fewer constraints, the values obtained by this procedure should be lower, which is indeed the case. Figure \ref{uncoupled} shows a similar floor structure of the landscape. However, in this case we do not have any theoretical result that tells us the value of its ground state; therefore, even though we find evidence of floor in high dimensions, it is hard tell how far away that floor value lies from the actual ground state. Since the order of growth of the function is still $N$, the ratio of the ground state to the value of the floor should also be the same as in the previous section, however we do not, at this point, have a proof for this.

\subsubsection{Teacher-student network: case of a known ground state at zero}

In this section we design a system for which we know where its global minimum lies. The idea for this experiment is inspired by the teacher-student network comparisons of \citet{NIPS1995_1072} and \citet{NIPS1996_1256}. It goes as follows: Split the MNIST training set into two parts. Using only the first half of the training data, train a network of two hidden layers, 784-500-300-10, with ReLU's at first two layers and soft-max at the last output layer. The teacher makes 211 mistakes in the test set of size 10,000. Use cross entropy loss and train the system with SGD. This gives the teacher network. Using the teacher network create new labels on the second half of the training set by replacing actual tags with the probabilities assigned by the outputs of the teacher network. 
\begin{table}[h!]
\caption{Student networks trained with SGD}
\begin{center}
\begin{tabular}{lcccc}
&\multicolumn{1}{c}{\bf Training cost} & \multicolumn{1}{c}{\bf Test cost} & \multicolumn{1}{c}{\bf Test error} &\multicolumn{1}{c}{\bf St.Dev.(test error)}
\\ \hline\\ 
784-50-50-10 student & 1.83e-02 & 1.69e-01 & 326 & 11.3\tabularnewline
784-250-150-10 student & 1.72e-02 & 1.31e-01 & 276 & 7.6\tabularnewline
784-500-300-10 student & 1.70e-02 & 1.25e-01 & 265 & 6.6\tabularnewline
784-1200-1200-10 student & 1.68e-02 & 1.18e-01 & 257 & 5.2\tabularnewline
\end{tabular} 
\end{center}
\end{table} 

If size of the student network is at least as big as the teacher network, it is guaranteed that zero cost value points exist. Moreover there are exponentially many of them: an exact copy of the network can be found, and appealing to the symmetries of the network one can permute all the weight connections without changing the cost. All student networks are trained with SGD, and it does not reach zero cost. The algorithm either gets stuck at a critical point or extremely slows down at a very flat area in the surface whose value is away from zero. We conjecture that this level is the floor for the student training surface. Also notice the different behavior of low dimensional networks that have bad critical points at which the floor is not well defined, which is again in contrast with the high dimensional networks. 

\begin{figure}[h]
  \begin{center}
\includegraphics[scale=0.44]{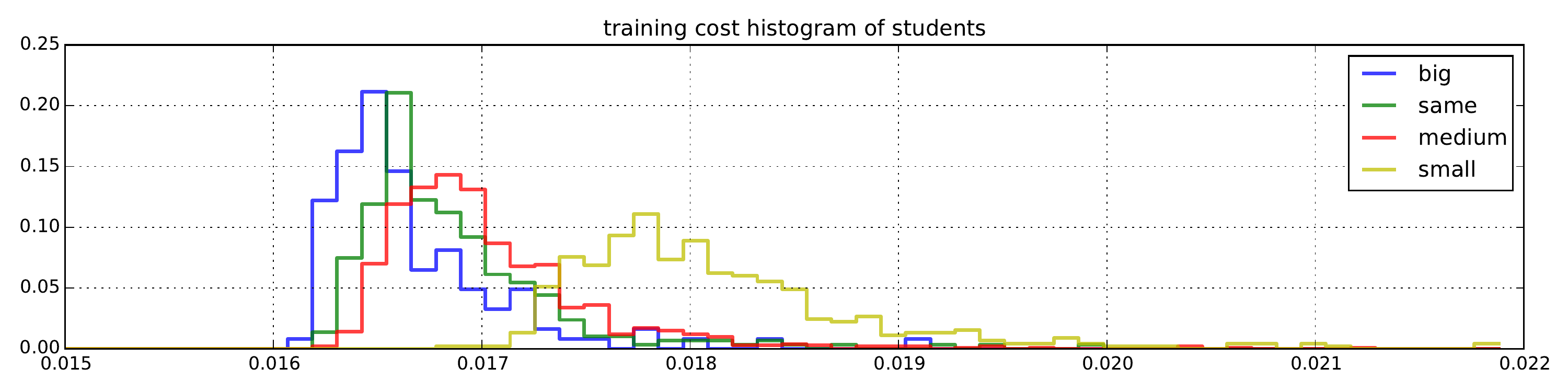}
\end{center}
\caption{Training cost values for different student network sizes. High dimensional surfaces exhibit a similar approach to the asymptotic floor value. The student networks that are at least of the same size as the teacher there is exponentially many possible zero values on the training surface, yet SGD can not locate them!}
\end{figure}

\begin{figure}[h!]
\includegraphics[width=11cm]{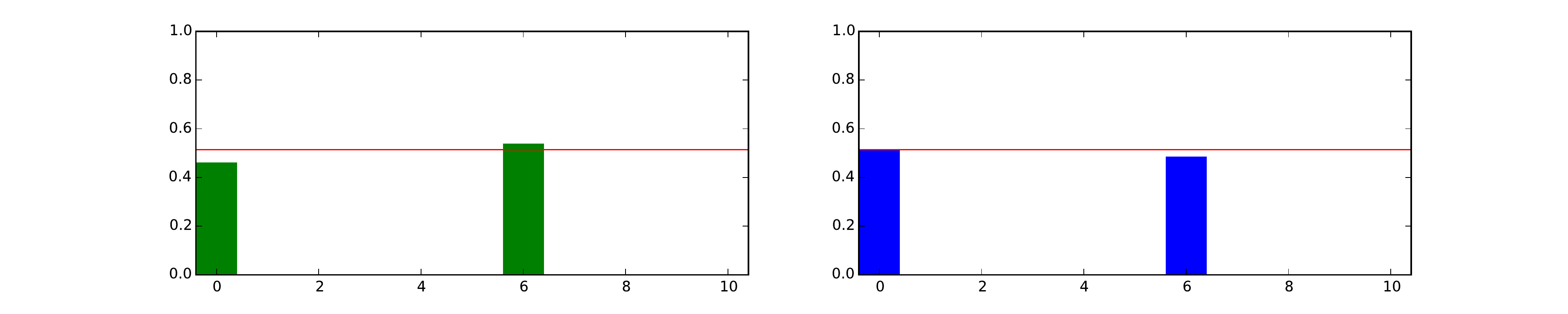}
\includegraphics[scale=0.221]{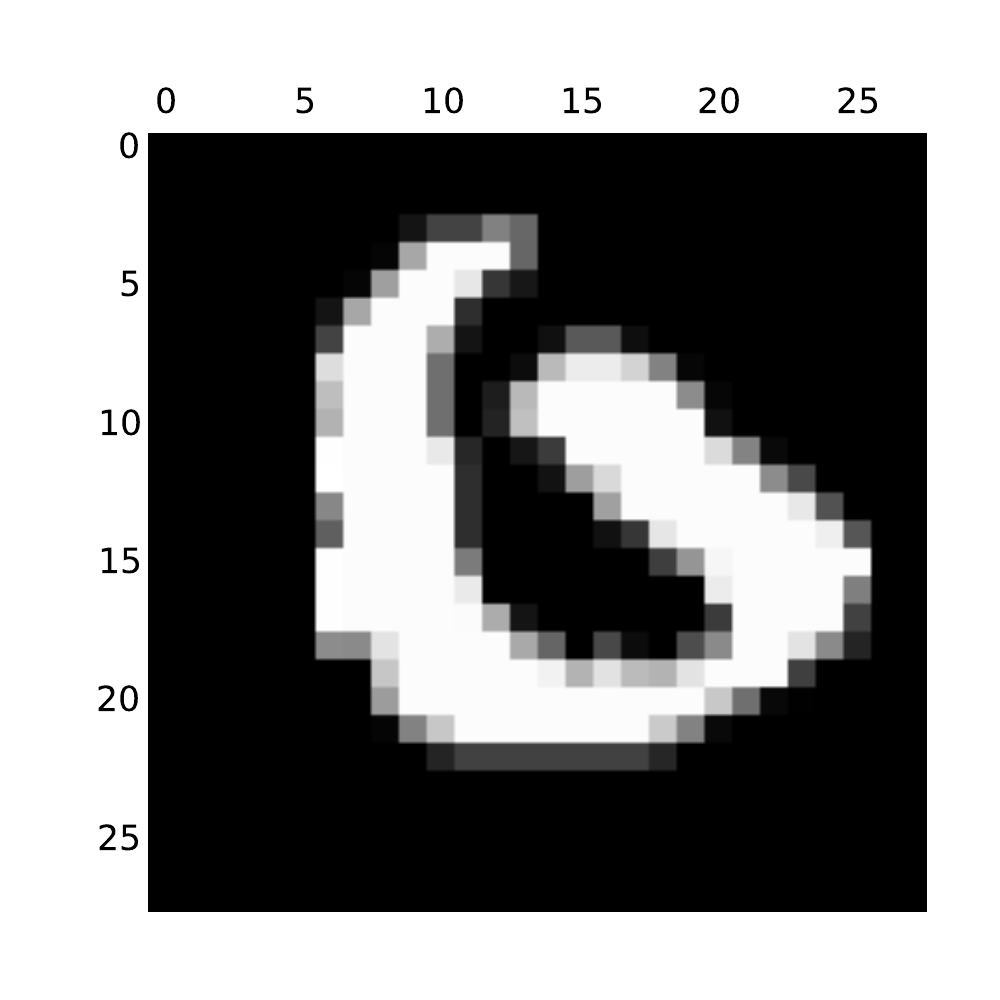}
\caption{One example of a medium sized student that correctly classifies a mistake of the teacher. The digit is seen on the right side. Histograms are probability outputs of the teacher (left) and the student (middle).} \label{vague}
\end{figure}

Data that students are trained on are not exact, they are partial in the sense that many labels are vague. For some characters, the teacher does not know the answer for sure. Figure \ref{vague} captures this vague response of the teacher which is learned by the student. But this ambiguity does not stop the teacher from teaching; instead, the teacher passes information with the ambiguity, which is some information by itself (this reminds us the Dark Knowledge at \citet{Hinton}). 

Recall that the teacher did not see the second half of the data during its training, but used it to teach its students. We look at its students to see how well they learned the things the teacher taught. It turns out, perhaps not surprisingly, that the ambiguity propagates. Many mistakes that the teacher makes in the second half of the data are also made by its students. There are cases that a student might judge more correctly. See Figure \ref{vague} for an example digit 0 that the teacher mistakenly thought was 6. Luckily, as an honest teacher, it did not train its student as if it were a firm 6. With the help of this extra information, the student was able to classify correctly the digit by a tiny margin. 

\subsection{Floor by gradient descent vs. stochastic gradient descent}

This section compares the two algorithms in terms of where they reach on their training surface.

\subsubsection{SGD for spin glass}

A spin glass field is created by a sum of smaller such fields:

\begin{equation}
\begin{split}
H_{N}(w) =\frac{1}{N}\sum_{i, j, k}^Nx_{ijk}w_{i}w_{j}w_{k} &=\sum_{p=1}^P \big( \frac{1}{N}\sum_{i, j, k}^Nx^p_{ijk}w_{i}w_{j}w_{k} \big) \label{ham2} \\
&=\frac{1}{N} \sum_{i, j, k}^N \big( \sum_{p=1}^P x^p_{ijk} \big) w_{i}w_{j}w_{k} \\
\end{split}
\end{equation}

Here $x^p_{ijk} \sim \text{ Gaussian}(0,\frac{1}{P})$ so that the resulting field of the sum is distributed as the one in equation \ref{ham}. Then at the $p^{th}$ step, the point on the sphere is updated in the negative direction of the gradient of the $p^{th}$ summand. This procedure is then the stochastic gradient descent with a minibatch of size 1. It is important to note that all summands are independent from each other, unlike the MNIST case. SGD still goes to the floor. The tiny difference in the values of SGD comes from the fact that the SGD slows down very fast and stops at the walls of a well. Once SGD slows down, one could restart GD from that point and reach the bottom of that well. 
\begin{figure}[h!]
\begin{center}
\includegraphics[scale=0.4]{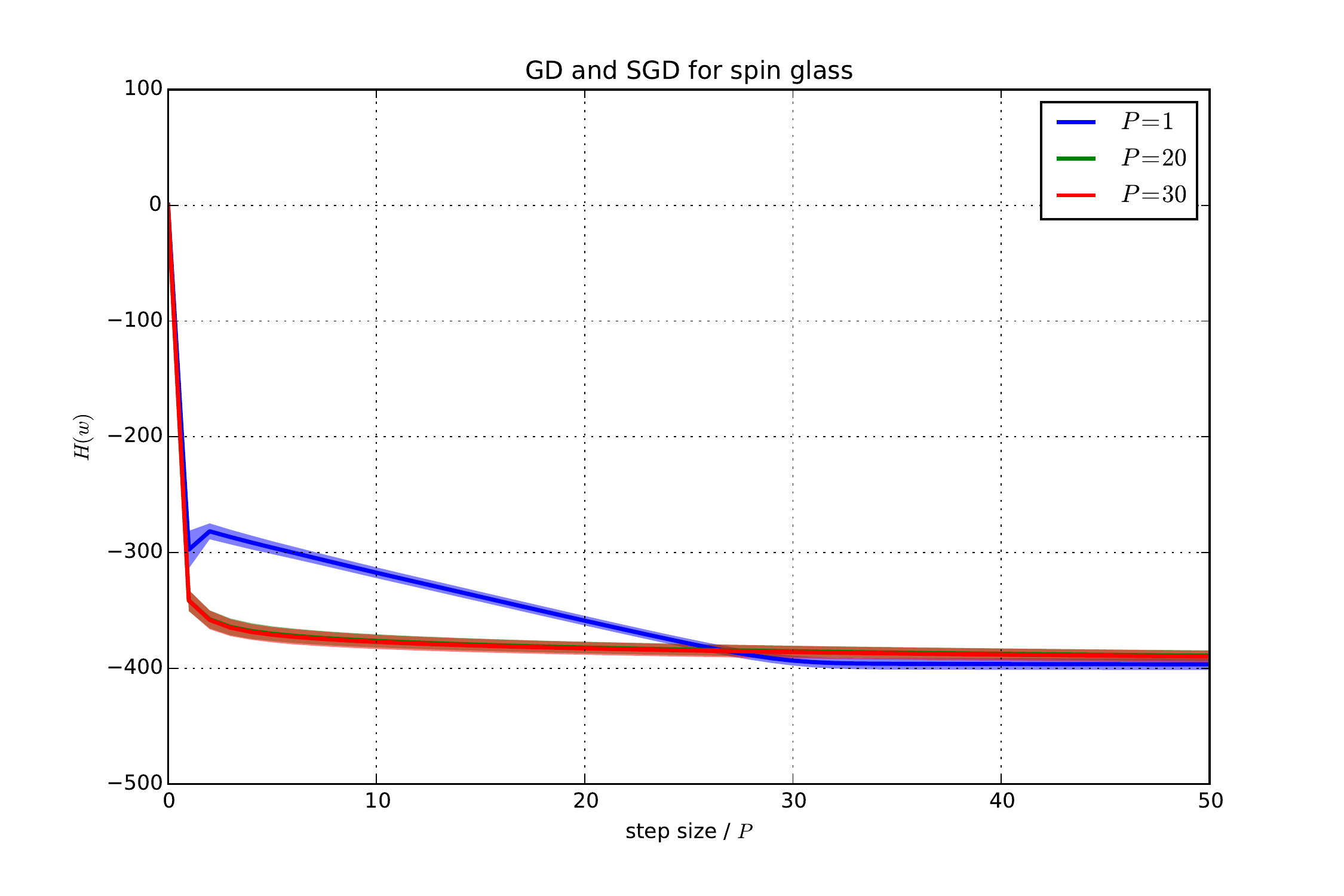}
\end{center}
\caption{Both methods reach the floor. Different $P$'s give eventually the same energy when measured at the same step-size-times-number-of-steps. $P=1$ corresponds to the gradient descent.} \label{GDvsSGDforSpins}
\end{figure}

\subsubsection{MNIST}
This experiment compares GD with SGD over full MNIST in a two layer network. One property that is attributed to SGD is that due to its noisy nature it is capable of escaping local minima at higher cost values. This would imply that GD would get stuck before SGD slows down. However it does not seem like this is the case at all. Within the same stepsize SGD and GD perfoms very similar on the training surface. Table \ref{GDvsSGDt} shows mean cost values and the difference in test errors.

\begin{figure}[h]
\begin{center}
\includegraphics[scale=0.4]{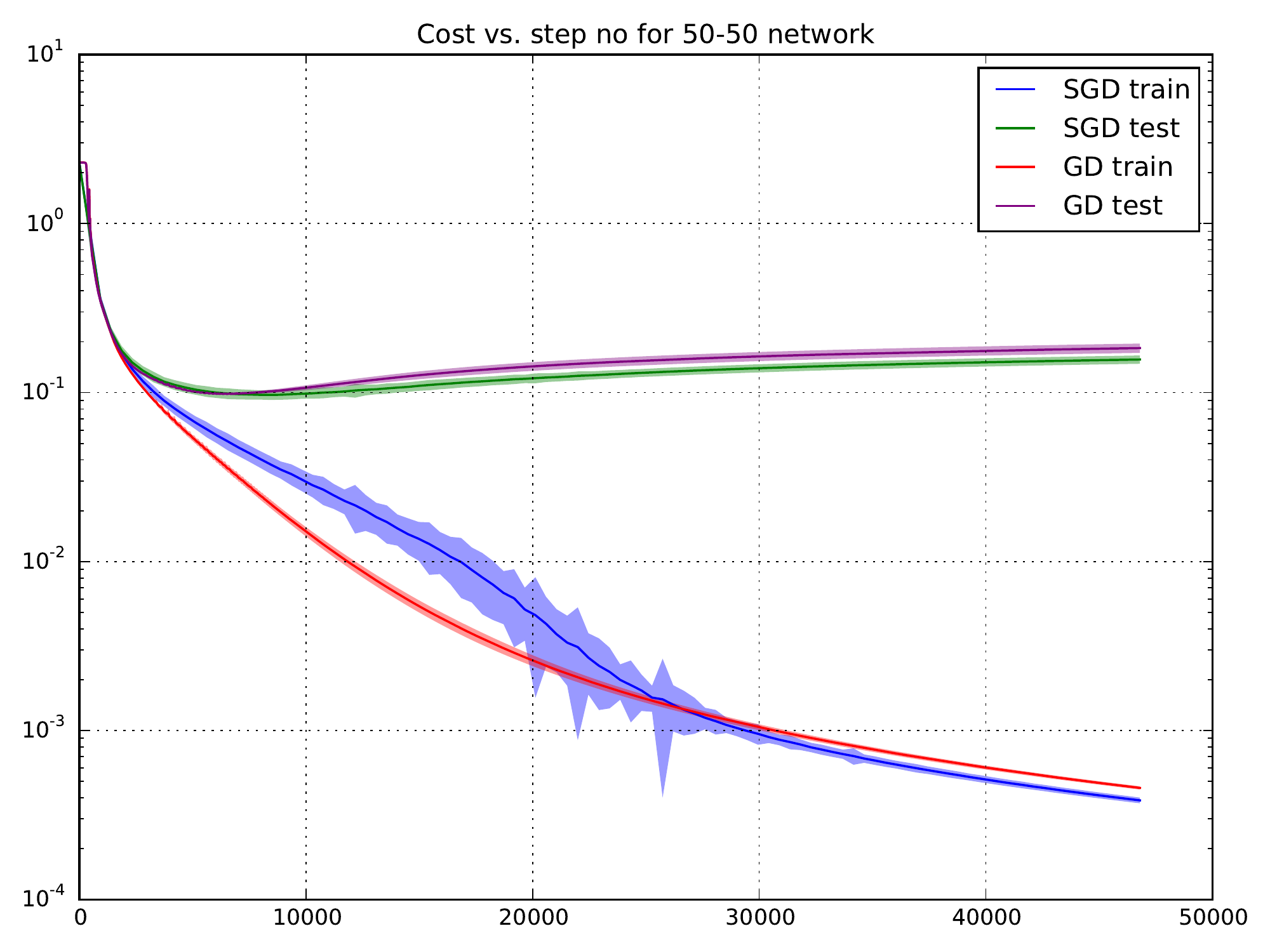}
\includegraphics[scale=0.4]{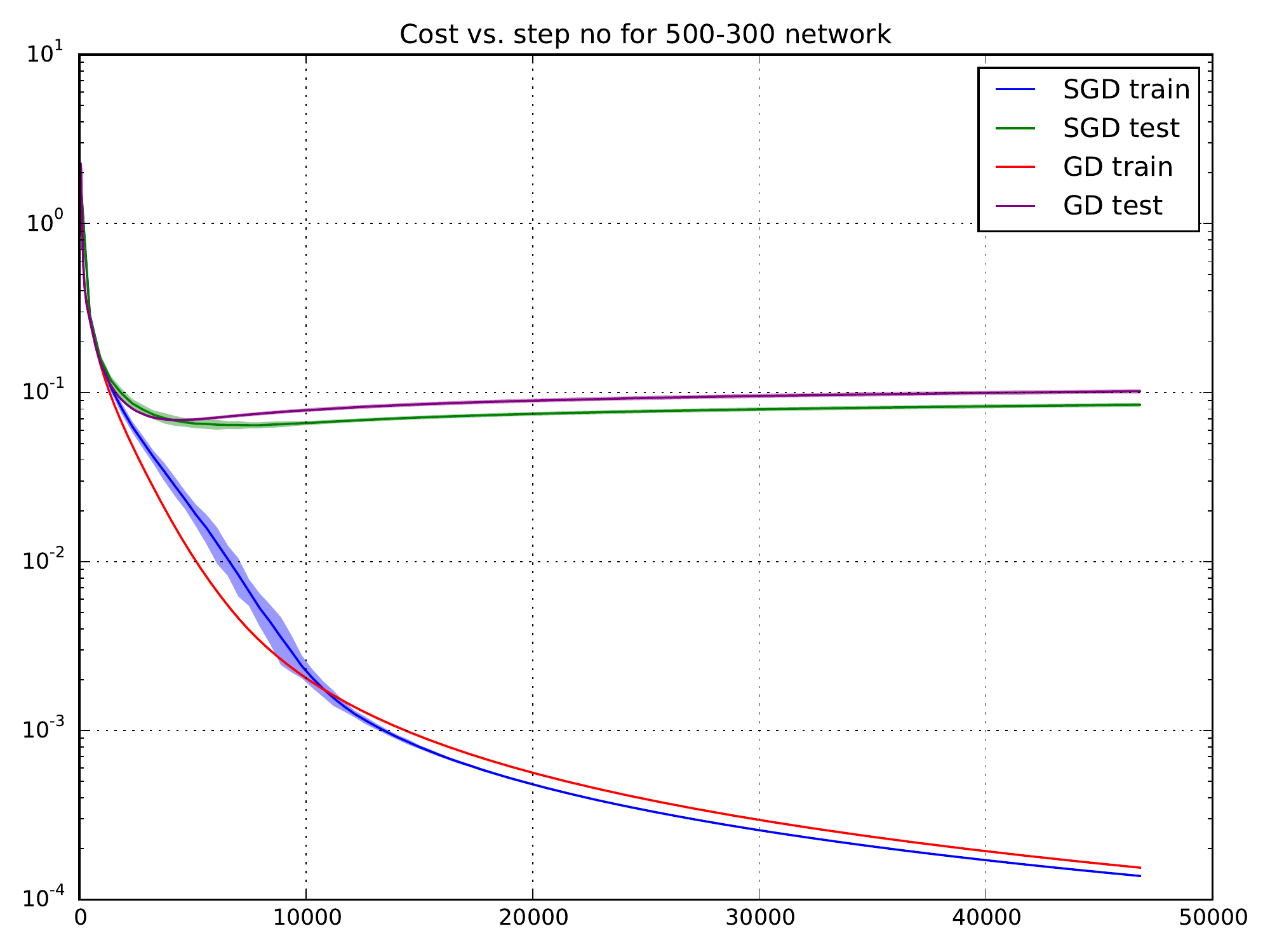}
\end{center}
\caption{Mean costs of GD and SGD experiments are given by the thin lines, and shady areas around the curve is the standard deviation around the mean. Vertical axis is in log scale. Decay of the function is not noticable in linear scale. Progress is very slow even in the log-scale. And the two curves follow each other tight.} \label{GDvsSGD}
\end{figure}

\begin{table}[t]
\caption{Averages of results for Gradient Descent and Stochastic Gradient Descent for MNIST } \label{GDvsSGDt}
\begin{center}
\begin{tabular}{lcccc}
&\multicolumn{1}{c}{\bf Training cost} & \multicolumn{1}{c}{\bf Test cost} & \multicolumn{1}{c}{\bf Test error} &\multicolumn{1}{c}{\bf St.Dev.test error}
\\ \hline\\
50-50 GD & 4.52e-04 & 2.02e-01 & 299 & 15.5\tabularnewline
50-50 SGD & 3.86e-04 & 1.57e-01 & 256 & 11.6\tabularnewline
500-300 GD & 1.55e-04 & 1.02e-01 & 194 & 6.9\tabularnewline
500-300 SGD & 1.38e-04 & 8.47e-02 & 174 & 5.3\tabularnewline
\end{tabular} 
\end{center}
\end{table}

\section{Conclusion}

High dimensional systems are typically considered as systems that come with their curses, however they also exhibit lots of symmetries which can be a blessing as observed in the first part of the simulations. One goal of the paper is to trigger theoretical research on non-convex surfaces on high dimensional domains. It is crucial to repeat here that we do not suggest a direct equivalency between spin glasses and deep networks; rather, we hint at a more general phenomenon that governs the two different cases. Finding similar properties in a variety of other problems might help us identify and quantify the properties of such complex systems. We hope further investigation in other optimization problems will lead to supporting conclusion that is in line with this research.

\subsubsection*{Acknowledgements}

We thank David Belius for valuable discussions, Taylan Cemgil and Atilla Y\i lmaz for valuable feedback, and reviewers for valuable suggestions. We thank the developers of Torch7 (\citet{Collobert_NIPSWORKSHOP_2011}) which we used for the spin glass simulations and Theano (\citet{Bastien-Theano-2012} and \citet{bergstra+al:2010-scipy}) which we used for MNIST experiments. We also gratefully acknowledge the support of NVIDIA Corporation with the donation of the Tesla K40 GPU used for part of this research.

\bibliography{cliffhanger}

\begin{thebibliography}{20}
\providecommand{\natexlab}[1]{#1}
\providecommand{\url}[1]{\texttt{#1}}
\expandafter\ifx\csname urlstyle\endcsname\relax
  \providecommand{\doi}[1]{doi: #1}\else
  \providecommand{\doi}{doi: \begingroup \urlstyle{rm}\Url}\fi

\bibitem[Agliari et~al.(2014)Agliari, Barra, Galluzzi, Tantari, and
  Tavani]{Agliari2014}
Agliari, Elena, Barra, Adriano, Galluzzi, Andrea, Tantari, Daniele, and Tavani,
  Flavia.
\newblock {A walk in the statistical mechanical formulation of neural
  networks}.
\newblock 1948:\penalty0 12, July 2014.

\bibitem[Auffinger \& {Ben Arous}(2013)Auffinger and {Ben
  Arous}]{Auffinger2013}
Auffinger, Antonio and {Ben Arous}, G\'{e}rard.
\newblock {Complexity of random smooth functions on the high-dimensional
  sphere}.
\newblock \emph{The Annals of Probability}, 41\penalty0 (6):\penalty0
  4214--4247, November 2013.
\newblock ISSN 0091-1798.
\newblock \doi{10.1214/13-AOP862}.

\bibitem[Auffinger \& Chen(2014)Auffinger and Chen]{Auffinger2014}
Auffinger, Antonio and Chen, Wei-kuo.
\newblock {Free Energy and Complexity of Spherical Bipartite Models}.
\newblock \emph{Journal of Statistical Physics}, 157\penalty0 (1):\penalty0
  40--59, July 2014.
\newblock ISSN 0022-4715.
\newblock \doi{10.1007/s10955-014-1073-0}.

\bibitem[Auffinger et~al.(2013)Auffinger, {Ben Arous}, and
  \v{C}ern\'{y}]{Auffinger2010}
Auffinger, Antonio, {Ben Arous}, G\'{e}rard, and \v{C}ern\'{y}, Ji\u{r}\'{\i}.
\newblock {Random Matrices and Complexity of Spin Glasses}.
\newblock \emph{Communications on Pure and Applied Mathematics}, 66\penalty0
  (2):\penalty0 165--201, February 2013.
\newblock ISSN 00103640.
\newblock \doi{10.1002/cpa.21422}.

\bibitem[Barra et~al.(2012)Barra, Genovese, Guerra, and
  Tantari]{1742-5468-2012-07-P07009}
Barra, Adriano, Genovese, Giuseppe, Guerra, Francesco, and Tantari, Daniele.
\newblock {How glassy are neural networks?}
\newblock \emph{Journal of Statistical Mechanics: Theory and Experiment},
  2012\penalty0 (07):\penalty0 P07009, 2012.

\bibitem[Bastien et~al.(2012)Bastien, Lamblin, Pascanu, Bergstra, Goodfellow,
  Bergeron, Bouchard, and Bengio]{Bastien-Theano-2012}
Bastien, Fr{\'{e}}d{\'{e}}ric, Lamblin, Pascal, Pascanu, Razvan, Bergstra,
  James, Goodfellow, Ian~J., Bergeron, Arnaud, Bouchard, Nicolas, and Bengio,
  Yoshua.
\newblock Theano: new features and speed improvements.
\newblock Deep Learning and Unsupervised Feature Learning NIPS 2012 Workshop,
  2012.

\bibitem[Bergstra et~al.(2010)Bergstra, Breuleux, Bastien, Lamblin, Pascanu,
  Desjardins, Turian, Warde-Farley, and Bengio]{bergstra+al:2010-scipy}
Bergstra, James, Breuleux, Olivier, Bastien, Fr{\'{e}}d{\'{e}}ric, Lamblin,
  Pascal, Pascanu, Razvan, Desjardins, Guillaume, Turian, Joseph, Warde-Farley,
  David, and Bengio, Yoshua.
\newblock Theano: a {CPU} and {GPU} math expression compiler.
\newblock In \emph{Proceedings of the Python for Scientific Computing
  Conference ({SciPy})}, June 2010.
\newblock Oral Presentation.

\bibitem[Choromanska et~al.(2014)Choromanska, Henaff, Mathieu, {Ben Arous}, and
  LeCun]{Choromanska}
Choromanska, Anna, Henaff, Mikael, Mathieu, Michael, {Ben Arous}, G\'{e}rard,
  and LeCun, Yann.
\newblock {The Loss Surface of Multilayer Networks}.
\newblock November 2014.

\bibitem[Collobert et~al.(2011)Collobert, Kavukcuoglu, and
  Farabet]{Collobert_NIPSWORKSHOP_2011}
Collobert, Ronan, Kavukcuoglu, Koray, and Farabet, Cl{\'{e}}ment.
\newblock Torch7: A matlab-like environment for machine learning.
\newblock In \emph{BigLearn, NIPS Workshop}, 2011.

\bibitem[Dauphin et~al.(2014)Dauphin, Pascanu, Gulcehre, Cho, Ganguli, and
  Bengio]{NIPS2014_5486}
Dauphin, Yann~N, Pascanu, Razvan, Gulcehre, Caglar, Cho, Kyunghyun, Ganguli,
  Surya, and Bengio, Yoshua.
\newblock {Identifying and attacking the saddle point problem in
  high-dimensional non-convex optimization}.
\newblock In Ghahramani, Z, Welling, M, Cortes, C, Lawrence, N~D, and
  Weinberger, K~Q (eds.), \emph{Advances in Neural Information Processing
  Systems 27}, pp.\  2933--2941. Curran Associates, Inc., June 2014.

\bibitem[Dean \& Majumdar(2008)Dean and Majumdar]{Dean2008}
Dean, David~S and Majumdar, Satya~N.
\newblock {Extreme value statistics of eigenvalues of Gaussian random
  matrices}.
\newblock \emph{Physical Review E}, 77\penalty0 (4):\penalty0 041108, April
  2008.
\newblock ISSN 1539-3755.
\newblock \doi{10.1103/PhysRevE.77.041108}.

\bibitem[Frieze \& Kannan(2008)Frieze and
  Kannan]{frieze_et_al:LIPIcs:2008:1752}
Frieze, Alan and Kannan, Ravi.
\newblock {A new approach to the planted clique problem}.
\newblock In Hariharan, Ramesh, Mukund, Madhavan, and Vinay, V (eds.),
  \emph{IARCS Annual Conference on Foundations of Software Technology and
  Theoretical Computer Science}, volume~2 of \emph{Leibniz International
  Proceedings in Informatics (LIPIcs)}, pp.\  187--198, Dagstuhl, Germany,
  2008. Schloss Dagstuhl--Leibniz-Zentrum fuer Informatik.
\newblock ISBN 978-3-939897-08-8.
\newblock \doi{http://dx.doi.org/10.4230/LIPIcs.FSTTCS.2008.1752}.

\bibitem[Fyodorov(2013)]{Fyodorov2013}
Fyodorov, Yan~V.
\newblock {High-Dimensional Random Fields and Random Matrix Theory}.
\newblock pp.\ ~40, July 2013.

\bibitem[Fyodorov \& {Le Doussal}(2013)Fyodorov and {Le
  Doussal}]{Fyodorov2013a}
Fyodorov, Yan~V and {Le Doussal}, Pierre.
\newblock {Topology Trivialization and Large Deviations for the Minimum in the
  Simplest Random Optimization}.
\newblock \emph{Journal of Statistical Physics}, 154\penalty0 (1-2):\penalty0
  466--490, September 2013.
\newblock ISSN 0022-4715.
\newblock \doi{10.1007/s10955-013-0838-1}.

\bibitem[Hinton et~al.()Hinton, Vinyals, and Dean]{Hinton}
Hinton, Geoffrey, Vinyals, Oriol, and Dean, Jeff.
\newblock {Dark Knowledge}.
\newblock \url{http://www.iro.umontreal.ca/~bengioy/cifar/NCAP2014-summerschool/slides/geoff_hinton_dark14.pdf}

\bibitem[Mehta et~al.({\natexlab{a}})Mehta, Hauenstein, Niemerg, Simm, and
  Stariolo]{Mehta}
Mehta, Dhagash, Hauenstein, Jonathan~D, Niemerg, Matthew, Simm, Nicholas~J, and
  Stariolo, Daniel~A.
\newblock {Energy Landscape of the Finite-Size Mean-field 2-Spin Spherical
  Model and Topology Trivialization}.
\newblock pp.\  1--9, {\natexlab{a}}.

\bibitem[Mehta et~al.({\natexlab{b}})Mehta, Stariolo, and Kastner]{Mehta2}
Mehta, Dhagash, Stariolo, Daniel~A, and Kastner, Michael.
\newblock {Energy Landscape of the Finite-Size Mean-field 3-Spin Spherical
  Model}.
\newblock pp.\  1--10, {\natexlab{b}}.

\bibitem[Saad et~al.(1996)Saad, Birmingham, and Solla]{NIPS1995_1072}
Saad, David, Birmingham, B, and Solla, Sara~A.
\newblock {Dynamics of On-Line Gradient Descent Learning for Multilayer Neural
  Networks}.
\newblock In Touretzky, D~S, Mozer, M~C, and Hasselmo, M~E (eds.),
  \emph{Advances in Neural Information Processing Systems 8}, pp.\  302--308.
  MIT Press, 1996.

\bibitem[Sherrington(2014)]{Sherrington2014}
Sherrington, David.
\newblock {Physics and Complexity: An Introduction}.
\newblock In Delitala, Marcello and {Ajmone Marsan}, Giulia (eds.),
  \emph{Managing Complexity, Reducing Perplexity}, volume~67 of \emph{Springer
  Proceedings in Mathematics \& Statistics}, pp.\  119--129. Springer
  International Publishing, Cham, 2014.
\newblock ISBN 978-3-319-03758-5.
\newblock \doi{10.1007/978-3-319-03759-2}.

\bibitem[West et~al.(1997)West, Saad, and Nabney]{NIPS1996_1256}
West, Ansgar H~L, Saad, David, and Nabney, Ian~T.
\newblock {The Learning Dynamcis of a Universal Approximator}.
\newblock In Mozer, M~C, Jordan, M~I, and Petsche, T (eds.), \emph{Advances in
  Neural Information Processing Systems 9}, pp.\  288--294. MIT Press, 1997.

\end{thebibliography}
\bibliographystyle{iclr2015}

\end{document}